\def\year{2017}\relax
\documentclass[letterpaper]{article}
\usepackage{aaai17}
\usepackage{times}
\usepackage{helvet}
\usepackage{courier}
\usepackage{amsmath}
\usepackage{amssymb}
\usepackage{graphicx}
\usepackage{multirow}
\usepackage{color}

\definecolor{zb_red}{RGB}{200, 0, 0}
\definecolor{zb_dred}{RGB}{100, 0, 0}
\definecolor{zb_lred}{RGB}{255, 102, 102}

\begin{document}
%
\title{BattRAE: Bidimensional Attention-Based Recursive Autoencoders for Learning Bilingual Phrase Embeddings}
\author{Biao Zhang$^{1}$, Deyi Xiong$^{2}$ \textmd{and} Jinsong Su$^{1}$\thanks{Corresponding author.}\\
	Xiamen University, Xiamen, China 361005$^{1}$ \\
	Soochow University, Suzhou, China 215006$^{2}$ \\
	{\tt zb@stu.xmu.edu.cn, dyxiong@suda.edu.cn, jssu@xmu.edu.cn} \\
}
\maketitle

\begin{abstract}
In this paper, we propose a bidimensional attention based recursive autoencoder (BattRAE) to integrate clues and source-target interactions at multiple levels of granularity into bilingual phrase representations. We employ recursive autoencoders to generate tree structures of phrases with embeddings at different levels of granularity (e.g., words, sub-phrases and phrases). Over these embeddings on the source and target side, we introduce a {\it bidimensional attention network} to learn their interactions encoded in a {\it bidimensional attention matrix}, from which we extract two soft attention weight distributions simultaneously. These weight distributions enable BattRAE to generate compositive phrase representations via convolution. Based on the learned phrase representations, we further use a bilinear neural model, trained via a max-margin method, to measure bilingual semantic similarity. To evaluate the effectiveness of BattRAE, we incorporate this semantic similarity as an additional feature into a state-of-the-art SMT system. Extensive experiments on NIST Chinese-English test sets show that our model achieves a substantial improvement of up to {\it 1.63} BLEU points on average over the baseline.
\end{abstract}

\section{Introduction} \label{introduction}

As one of the most important components in statistical machine translation (SMT), translation model measures the translation faithfulness of a hypothesis to a source fragment~\cite{och03,Koehn:2003:SPT:1073445.1073462,Chiang:2007:HPT:1268656.1268659}. Conventional translation models extract a huge number of {\it bilingual phrases} with conditional translation probabilities and lexical weights~\cite{Koehn:2003:SPT:1073445.1073462}. Due to the heavy reliance of the calculation of these probabilities and weights on surface forms of bilingual phrases, traditional translation models often suffer from the problem of data sparsity. This leads researchers to investigate methods that learn underlying semantic representations of phrases using neural networks~\cite{gao-EtAl:2014:P14-1,zhang-EtAl:2014:P14-11,cho-EtAl:2014:EMNLP2014,su-EtAl:2015:EMNLP2}.

\begin{table}[t]
\centering
\small
\begin{tabular}{c|c}
\hline
{\bf Src (Ref)} & {\bf Tgt} \\
\hline
\hline
{\it \textbf{\textit{shijie}} geda chengshi} & \multirow{2}{*}{\it in other major cities} \\
{\it (major cities in the world)} & \\
\hline
{\it dui \textbf{\textit{jingji xuezhe}}} & \multirow{2}{*}{\it to \textbf{\textit{economists}}} \\
{\it (to the economists)} & \\
\hline
\end{tabular}
\caption{\label{tab_intuitive_c2e} Examples of bilingual phrases ((romanized) Chinese-English) from our translation model. The important words or phrases are highlighted in bold. {\bf Src} = source, {\bf Tgt} = target, {\bf Ref} = literal translation of source phrase.}
\end{table}

Typically, these neural models learn bilingual phrase embeddings in a way that embeddings of source and corresponding target phrases are optimized to be close as much as possible in a continuous space. In spite of their success, they either explore {\bf clues} (linguistic items from contexts) only at a single level of granularity or capture {\bf interactions} (alignments between source and target items) only at the same level of granularity to learn bilingual phrase embeddings. 

We believe that clues and interactions from a single level of granularity are not adequate to measure underlying semantic similarity of bilingual phrases due to the high language divergence. Take the Chinese-English translation pairs in Table \ref{tab_intuitive_c2e} as examples. At the word level of granularity, we can easily recognize that the translation of the first instance is not faithful as Chinese word ``{\it shijie}'' ({\it world}) is not translated at all. While in the second instance, semantic judgment at the word level is not sufficient as there is no translation for single Chinese word ``{\it jingji}'' ({\it economy}) or ``{\it xuezhe}'' ({\it scholar}). We have to elevate the calculation of semantic similarity to a higher sub-phrase level: ``{\it jingji xuezhe}'' vs. ``{\it economists}''. This suggests that clues and interactions between the source and target side at multiple levels of granularity should be explored to measure semantic similarity of bilingual phrases.

\begin{figure}[t]
\centering
\includegraphics[scale=0.60]{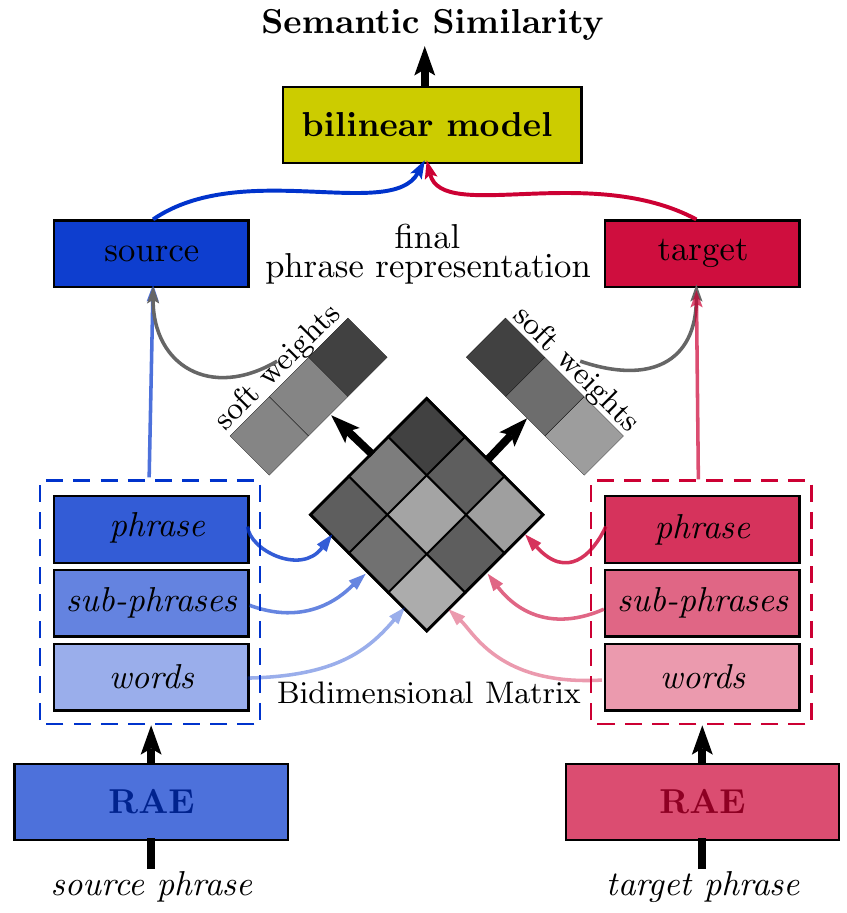}
\caption{\label{fig_overall_arche} Overall architecture for the proposed BattRAE model. We use blue and red color to indicate the source- and target-related representations or structures respectively. The gray colors indicate real values in the bidimensional mechanism.}
\end{figure}

In order to capture multi-level clues and interactions, we propose a bidimensional attention based recursive autoencoder (BattRAE). It learns bilingual phrase embeddings according to the strengths of interactions between linguistic items at different levels of granularity (i.e., words, sub-phrases and entire phrases) on the source side and those on the target side. The philosophy behind BattRAE is twofold: 1) Phrase embeddings are learned from weighted clues at different levels of granularity; 2) The weights of clues are calculated according to the alignments of linguistic items at different levels of granularity between the source and target side. We introduce a {\it bidimensional attention network} to learn the strengths of these alignments. Figure \ref{fig_overall_arche} illustrates the overall architecture of BattRAE model. Specifically,
\begin{itemize}
\item 
First, we adopt recursive autoencoders to generate hierarchical structures of source and target phrases separately. At the same time, we can also obtain embeddings of multiple levels of granularity, i.e., words, sub-phrases and entire phrases from the generated structures. (see Section \ref{rae})
\item
Second, BattRAE projects the representations of linguistic items at different levels of granularity onto an attention space, upon which the alignment strengths of linguistic items from the source and target side are calculated by estimating how well they semantically match. These alignment scores are stored in a bidimensional attention matrix. Over this matrix, we perform row (column)-wise summation and softmax operations to generate attention weights on the source (target) side. The final phrase representations are computed as the weighted sum of their initial embeddings using these attention weights. (see Section\ref{biattention})
\item
Finally, BattRAE projects the bilingual phrase representations onto a common semantic space, and uses a bilinear model to measure their semantic similarity. (see Section \ref{semantic})
\end{itemize}
We train the BattRAE model with a max-margin method, which maximizes the semantic similarity of translation equivalents and minimizes that of non-translation pairs (see Section \ref{objection}). 

In order to verify the effectiveness of BattRAE in learning bilingual phrase representations, we incorporate the learned semantic similarity of bilingual phrases as a new feature into SMT for translation selection. We conduct experiments with a state-of-the-art SMT system on large-scale training data. Results on the NIST 2006 and 2008 datasets show that BattRAE achieves significant improvements over baseline methods. Further analysis on the bidimensional attention matrix reveals that BattRAE is able to detect semantically related parts of bilingual phrases and assign higher weights to these parts for constructing final bilingual phrase embeddings than those not semantically related.

\section{Learning Embeddings at Different Levels of Granularity}\label{rae}

We use recursive autoencoders (RAE) to learn initial embeddings at different levels of granularity for our model. Combining two children vectors from the bottom up recursively, RAE is able to generate low-dimensional vector representations for variable-sized sequences. The recursion procedure usually consists of two neural operations: {\it composition} and {\it reconstruction}.

{\it Composition:} Typically, the input to RAE is a list of ordered words in a phrase $(x_1, x_2, x_3)$, each of which is embedded into a $d$-dimensional continuous vector.\footnote{Generally, all these word vectors are stacked into a word embedding matrix $L \in \mathbb{R}^{d \times |V|}$, where $|V|$ is the size of the vocabulary.} In each recursion, RAE selects two neighboring children (e.g. $c_1 = x_1$ and $c_2 = x_2$) via some selection criterion, and then compose them into a parent embedding $y_1$, which can be computed as follows:
\begin{equation}
y_1 = f(W^{(1)}[c_1;c_2] + b^{(1)})
\end{equation}
where $[c_1; c_2] \in \mathbb{R}^{2d}$ is the concatenation of $c_1$ and $c_2$, $W^{(1)} \in \mathbb{R}^{d \times 2d}$ is a parameter matrix, $b^{(1)} \in \mathbb{R}^{d}$ is a bias term, and $f$ is an element-wise activation function such as $tanh(\cdot)$, which is used in our experiments.

{\it Reconstruction:} After the composition, we obtain the representation for the parent $y_1$ which is also a $d$-dimensional vector. In order to measure how well the parent $y_1$ represents its children, we reconstruct the original child nodes via a reconstruction layer:
\begin{equation}
\lbrack c^{\prime}_1; c^{\prime}_2 \rbrack = f(W^{(2)}y_1 + b^{(2)})  \label{parent-to-children}
\end{equation}
here $c^{\prime}_1$ and $c^{\prime}_2$ are the reconstructed children, $W^{(2)} \in \mathbb{R}^{2d \times d}$ and $b^{(2)} \in \mathbb{R}^{2d}$. The minimum Euclidean distance between $[c^{\prime}_1; c^{\prime}_2]$ and $[c_1; c_2]$ is usually used as the selection criterion during composition.

These two standard processes form the basic procedure of RAE, which repeat until the embedding of the entire phrase is generated. In addition to phrase embeddings, RAE also constructs a binary tree. The structure of the tree is determined by the used selection criterion in composition. As generating the optimal binary tree for a phrase is usually intractable, we employ a greedy algorithm \cite{socher-EtAl:2011:EMNLP} based on the following reconstruction error:
\begin{equation}
E_{rec}(\mathbf{x}) = \sum\limits_{y \in T(\mathbf{x})} \frac{1}{2} \parallel [c_1; c_2]_y - [c^{\prime}_1; c^{\prime}_2]_y \parallel^2 \label{reconstruction-error}
\end{equation}
Parameters $W^{(1)}$ and $W^{(2)}$ are thereby learned to minimize the sum of reconstruction errors at each intermediate node $y$ in the binary tree $T(\mathbf{x})$. For more details, we refer the readers to~\cite{socher-EtAl:2011:EMNLP}.

Given an binary tree learned by RAE, we regard each level of the tree as {\bf a level of granularity}. In this way, we can use RAE to produce embeddings of linguistic expressions at different levels of granularity. Unfortunately, RAE is unable to synthesize embeddings across different levels of granularity, which will be discussed in the next section.

Additionally, as illustrated in Figure \ref{fig_overall_arche}, RAEs for the source and target language are learned separately. In our model, we assume that phrase embeddings for different languages are from different semantic spaces. To make this clear, we denote dimensions of source and target phrase embeddings as $d_{s}$ and $d_{t}$ respectively. 

\section{Bidimensional Attention-Based Recursive Autoencoders}

In this section, we present the proposed BattRAE model. We first elaborate the bidimensional attention network, and then the semantic similarity model built on phrase embeddings learned with the attention network. Finally, we introduce the objective function and training procedure. 

\subsection{Bidimensional Attention Network}\label{biattention}

\begin{figure}[t]
\centering
\includegraphics[scale=0.60]{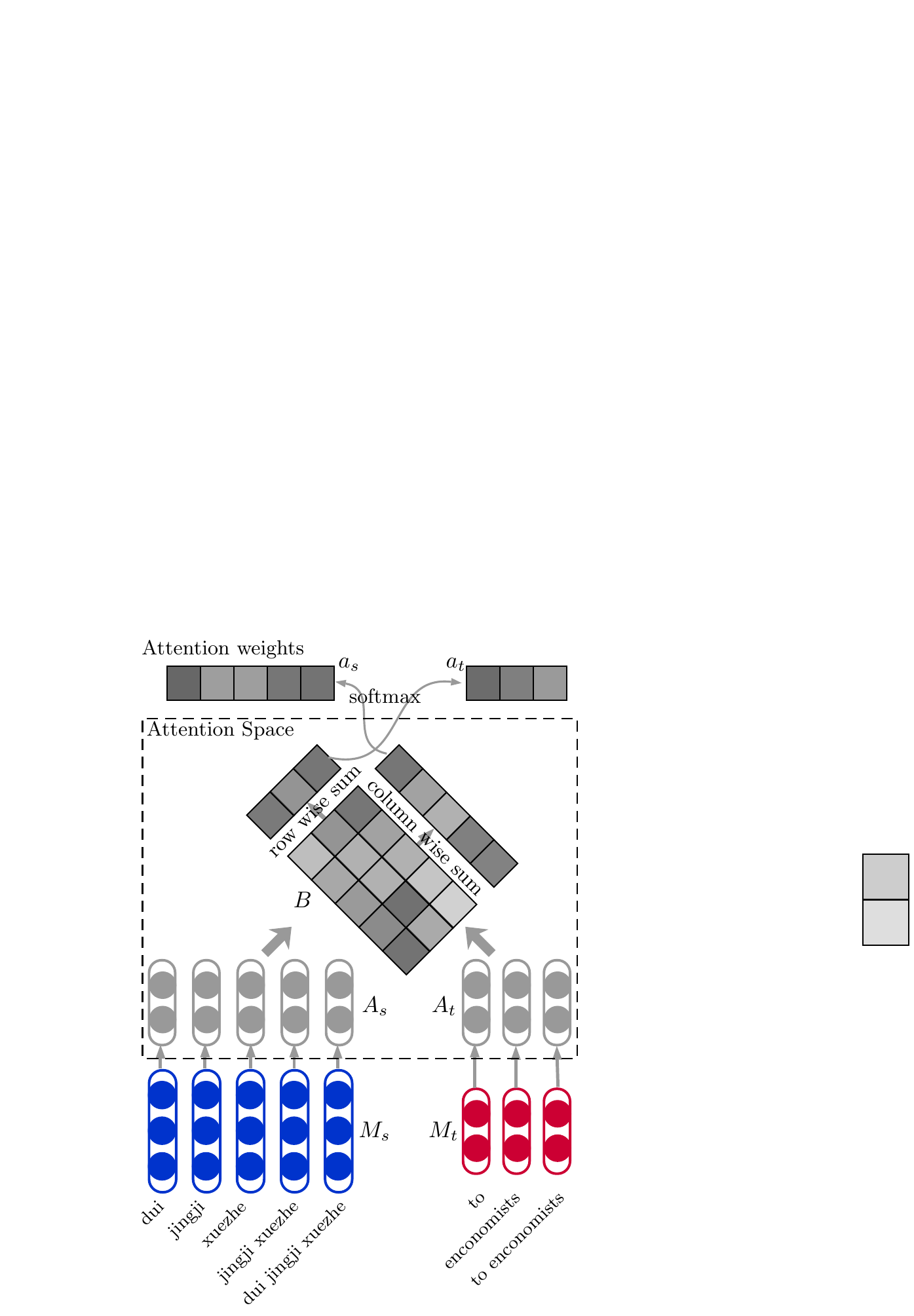}
\caption{\label{fig_biattention} An illustration of the bidimensional attention network in BattRAE model. The gray circles represent the attention space. We use subscript $s$ and $t$ to indicate the source and target respectively.}
\end{figure}

As mentioned in Section \ref{introduction}, we would like to incorporate clues and interactions at multiple levels of granularity into phrase embeddings and further into the semantic similarity model of bilingual phrases. The clues are encoded in multi-level embeddings learned by RAEs. The interactions between linguistic items on the source and target side can be measured by how well they semantically match. In order to jointly model clues and interactions at multiple levels of granularity, we propose the {\it bidimensional attention network}, which is illustrated in Figure \ref{fig_biattention}.

We take the bilingual phrase (``{\it dui jingji xuezhe}'', ``{\it to economists}'') in Table \ref{tab_intuitive_c2e} as an example. Let's suppose that their phrase structures learned by RAE are ``({\it dui}, ({\it jingji}, {\it xuezhe}))'' and ``({\it to}, {\it economists})'' respectively. We perform a postorder traversal on these structures to extract the embeddings of words, sub-phrases and the entire source/target phrase. We treat each embedding as a column and put them together to form a matrix $M_s \in \mathbb{R}^{d_s \times n_s}$ on the source side and $M_t \in \mathbb{R}^{d_t \times n_t}$ on the target side. Here, $n_s$=$5$ and $M_s$ contains embeddings from linguistic items (``{\it dui}'', ``{\it jingji}'', ``{\it xuezhe}'', ``{\it jingji xuezhe}'', ``{\it dui jingji xuezhe}'') at different levels of granularity.  Similarly, $n_t$=$3$ and $M_t$ contains embeddings of (``{\it to}'', ``{\it economists}'', ``{\it to economists}''). $M_s$ and $M_t$ form the input layer for the bidimensional attention network.

We further stack an attention layer upon the input matrices to project the embeddings from $M_s$ and $M_t$ onto a common attention space as follows (see the gray circles in Figure \ref{fig_biattention}):
\begin{equation}
A_s = f(W^{(3)} M_s + b^{A}_{[:]}) \label{att_s}
\end{equation}
\begin{equation}
A_t = f(W^{(4)} M_t + b^{A}_{[:]}) \label{att_t}
\end{equation}
where $W^{(3)} \in \mathbb{R}^{d_a \times d_s}$, $W^{(4)} \in \mathbb{R}^{d_a \times d_t}$ are transformation matrices, $b^{A} \in \mathbb{R}^{d_a}$ is the bias term, and $d_a$ is the dimensionality of the attention space. The subscript $[:]$ indicates a {\it broadcasting} operation. Note that we use different transformation matrices but share the same bias term for $A_s$ and $A_t$. This will force our model to learn to encode attention semantics into the two transformation matrices, rather than the bias term.

On this attention space, each embedding from the source side is able to interact with all embeddings from the target side and vice versa. The strength of such an interaction can be measured by a semantically matching score, which is calculated via the following equation:
\begin{equation} \label{attention}
B_{i,j} = g(A_{s,i}^T A_{t,j})
\end{equation}
where $B_{i,j} \in \mathbb{R}$ is the score that measures how well the $i$-th column embedding in $A_s$ semantically matches the $j$-th column embedding in $A_t$, and $g(\cdot)$ is a non-linear function, e.g., the sigmoid function used in this paper. All matching scores form a matrix $B \in \mathbb{R}^{n_s \times n_t}$, which we call the {\it bidimensional attention matrix}. Intuitively, this matrix is a result of handshakes between the source and target phrase at multiple levels of granularity.

Given the bidimensional attention matrix, our next interest lies in how important an embedding at a specific level of granularity is to the semantic similarity between the corresponding source and target phrase. As each embedding interacts all embeddings on the other side, its importance can be measured as the summation of the strengths of all these interactions, i.e., matching scores computed in Eq. (\ref{attention}). This can be done via a row/column-wise summation operation over the bidimensional attention matrix as follows.
\begin{equation}
\tilde{a}_{s, i} = \sum_{j} B_{i,j}, ~~
\tilde{a}_{t, j} = \sum_{i} B_{i,j}
\end{equation}
where $\tilde{a}_s \in \mathbb{R}^{n_s}$ and $\tilde{a}_t \in \mathbb{R}^{n_t}$ are the matching score vectors.

Because the length of a phrase is uncertain, we apply a Softmax operation on $\tilde{a}_s$ and $\tilde{a}_t$ to keep their values  at the same magnitude: $a_s = Softmax(\tilde{a}_s), a_t = Softmax(\tilde{a}_t)$. This forces $a_s$ and $a_t$ to become real-valued distributions on the attention space. We call them {\it attention weights} (see Figure \ref{fig_biattention}). An important feature of this attention mechanism is that it naturally deals with variable-length bilingual inputs (as we do not impose any length constrains on $n_s$ and $n_t$ at all).

To obtain final bilingual phrase representations, we convolute the embeddings in phrase structures with the computed attention weights:
\begin{equation}
p_s = \sum_{i} a_{s,i} M_{s, i}, ~~p_t = \sum_{j} a_{t,j} M_{t, j}
\end{equation}
This ensures that the generated phrase representations encode weighted clues and interactions at multiple levels of granularity between the source and target phrase. Notice that $p_s \in \mathbb{R}^{d_s}$ and $p_t \in \mathbb{R}^{d_t}$ still locate in their language-specific vector space. 

\subsection{Semantic Similarity}\label{semantic}

To measure the semantic similarity for a bilingual phrase, we first transform the learned bilingual phrase representations $p_s$ and $p_t$ into a common semantic space through a non-linear projection as follows:
\begin{equation}
s_s = f(W^{(5)} p_s + b^{s})
\end{equation}
\begin{equation}
s_t = f(W^{(6)} p_t + b^{s})
\end{equation}
where $W^{(5)} \in \mathbb{R}^{d_{sem} \times d_s}$, $W^{(6)} \in \mathbb{R}^{d_{sem} \times d_t}$ and $b^{s} \in \mathbb{R}^{d_{sem}}$ are the parameters. Similar to the transformation in Eq. (\ref{att_s}) and (\ref{att_t}), we share the same bias term for both $s_s$ and $s_t$.

We then use a bilingual model to compute the semantic similarity score as follows:
\begin{equation}
s(f,e) = s_s^T S s_t
\end{equation}
where $f$ and $e$ is the source and target phrase respectively, and $s(\cdot, \cdot)$ represents the semantic similarity function. $S \in \mathbb{R}^{d_{sem} \times d_{sem}}$ is a squared matrix of parameters to be learned. We choose this model because that the matrix $S$ actually represents an interaction between $s_s$ and $s_t$, which is desired for our purpose.

\subsection{Objective and Training}\label{objection}

There are two kinds of errors involved in our objective function: {\it reconstruction error} (see Eq. (\ref{reconstruction-error})) and {\it semantic error}. The latter error measures how well a source phrase semantically match its counterpart target phrase. We employ a max-margin method to estimate this semantic error. Given a training instance $(f, e)$ with negative samples $(f^{-}, e^{-})$, we define the following ranking-based error:
\begin{align}
E_{sem}(f, e) & =  max(0, 1+s(f,e^{-}) - s(f,e)) \notag \\
 + &max(0, 1+s(f^{-}, e) - s(f,e))
\end{align}
Intuitively, minimizing this error will maximize the semantic similarity of the correct translation pair $(f, e)$ and minimize (up to a margin) the similarity of negative translation pairs $(f^-, e)$ and $(f, e^-)$. In order to generate the negative samples, we replace words in a correct translation pair with random words, which is similar to the sampling method used by Zhang et al.~\shortcite{zhang-EtAl:2014:P14-11}.

For each training instance $(f,e)$, the joint objective of BattRAE is defined as follows:
\begin{align}
J(\theta) = \alpha E_{rec}(f,e) + \beta & E_{sem}(f,e) + R(\theta)
\end{align}
where $E_{rec}(f,e) = E_{rec}(f) + E_{rec}(e)$, parameters $\alpha$ and $\beta$ ($\alpha + \beta = 1$) are used to balance the preference between the two errors, and $R(\theta)$ is the regularization term. We divide the parameters $\theta$ into four different groups\footnote{The subscript $s$ and $t$ are used to denote the source and the target language.}:
\begin{enumerate}
\item $\theta_L:$ the word embedding matrices $L_s$ and $L_t$;
\item $\theta_{rec}:$ the parameters for RAE $W^{(1)}_s$, $W^{(1)}_t$, $W^{(2)}_s$, $W^{(2)}_t$ and $b^{(1)}_s$, $b^{(1)}_t$, $b^{(2)}_s$, $b^{(2)}_t$;
\item $\theta_{att}:$ the parameters for the projection of the input matrices onto the attention space $W^{(3)},W^{(4)}$ and $b^{A}$;
\item $\theta_{sem}:$ the parameters for semantic similarity computation $W^{(5)},W^{(6)},S$ and $b^s$;
\end{enumerate}
And each parameter group is regularized with a unique weight:
\begin{align}
\scriptstyle 
R(\theta) = \frac{\lambda_{L}}{2} \|\theta_{L}\|^2 + \frac{\lambda_{rec}}{2} \|\theta_{rec}\|^2 + \frac{\lambda_{att}}{2}\|\theta_{att}\|^2
+ \frac{\lambda_{sem}}{2}\|\theta_{sem}\|^2 \label{regularizer-defination}
\end{align}
where $\lambda_{*}$ are our hyperparameters.

To optimize these parameters, we apply the L-BFGS algorithm which requires two conditions: {\it parameter initialization} and {\it gradient calculation}.

{\it Parameter Initialization:} We randomly initialize $\theta_{rec}$, $\theta_{att}$ and $\theta_{sem}$ according to a normal distribution ($\mu$=$0$,$\sigma$=$0.01$). With respect to the word embeddings $\theta_L$, we use the toolkit Word2Vec\footnote{https://code.google.com/p/word2vec/} to pretrain them on a large scale unlabeled data. All these parameters will be further fine-tuned in our BattRAE model.

{\it Gradient Calculation:} We compute the partial gradient for parameter $\theta_k$ as follows:
\begin{align}
\frac{\partial J}{\partial \theta_k} = \frac{\partial E_{rec}(f,e)}{\partial \theta_k} + \frac{\partial E_{sem}(f,e)}{\partial \theta_k} +\lambda_k \theta_k
\end{align}
This gradient is fed into the toolkit libLBFGS\footnote{http://www.chokkan.org/software/liblbfgs/} for parameter updating in our practical implementation.

\section{Experiment}

In order to examine the effectiveness of BattRAE in learning bilingual phrase embeddings, we carried out large scale experiments on NIST Chinese-English translation tasks.\footnote{Source code is available at https://github.com/DeepLearnXMU/BattRAE.}

\subsection{Setup}

Our parallel corpus consists of 1.25M sentence pairs extracted from LDC corpora\footnote{This includes LDC2002E18, LDC2003E07, LDC2003E14, Hansards portion of LDC2004T07, LDC2004T08 and LDC2005T06.}, with 27.9M Chinese words and 34.5M English words respectively. We trained a 5-gram language model on the Xinhua portion of the GIGAWORD corpus (247.6M English words) using \textit{SRILM} Toolkit\footnote{http://www.speech.sri.com/projects/srilm/download.html} with modified Kneser-Ney Smoothing. We used the NIST MT05 data set as the development set, and the NIST MT06/MT08 datasets as the test sets. We used minimum error rate training~\cite{och03} to optimize the weights of our translation system. We used case-insensitive BLEU-4 metric~\cite{PapineniEtAl2002} to evaluate translation quality and  performed the paired bootstrap sampling~\cite{koehn04} for significance test.

In order to obtain high-quality bilingual phrases to train the BattRAE model, we used forced decoding~\cite{wuebker-mauser-ney:2010:ACL} (but without the leaving-one-out) on the above parallel corpus, and collected 2.8M phrase pairs. From these pairs, we further extracted 34K bilingual phrases as our development data to optimize all hyper-parameters using {\it random search}~\cite{Bergstra+Bengio-2012}. Finally, we set $d_s$=$d_t$=$d_a$=$d_{sem}$=$50$, $\alpha$=$0.125$ (such that, $\beta$=$0.875$), $\lambda_L$=$1e^{-5}$, $\lambda_{rec}$=$\lambda_{att}$=$1e^{-4}$ and $\lambda_{sem}$=$1e^{-3}$ according to experiments on the development data. Additionally, we set the maximum number of iterations in the L-BFGS algorithm to 100.

\begin{table}[t]
\begin{center}
{ \small
\begin{tabular}{c|l|l|l}
\hline
\multicolumn{1}{c|}{\bf Method} &
\multicolumn{1}{|l|}{\bf MT06 } &
\multicolumn{1}{|l|}{\bf MT08 } &
\multicolumn{1}{|l}{\bf AVG} \\
\hline
\hline
{\it Baseline} & 31.55 & 23.66 & 27.61 \\
\hline
{\it BRAE } & 32.29 & 24.30 & 28.30 \\
{\it BCorrRAE} & 32.62 & 24.89 & 28.76 \\
\hline
{\it BattRAE} & {\bf 33.19}$^{\Uparrow**\uparrow}$ & {\bf 25.29}$^{\Uparrow**}$ & {\bf 29.24} \\
\hline
\end{tabular}
}
\end{center}
\caption{\label{overall-performance} Experiment results on the MT 06/08 test sets. {\bf AVG} = average BLEU scores for test sets. We highlight the best result in bold. ``$\Uparrow$'': significantly better than {\it Baseline} ($p$ $<$ $0.01$); ``$**$'': significantly better than {\it BRAE} ($p$ $<$ $0.01$); ``$\uparrow$'': significantly better than {\it BCorrRAE} ($p$ $<$ $0.05$). }
\end{table}

\subsection{Translation Performance}

\begin{table*}[t]
\centering
\small
\begin{tabular}{c|c|c|c|c}
\hline
\multirow{2}{*}{\bf Type} & \multicolumn{2}{c|}{\bf Phrase Structure} & \multicolumn{2}{|c}{\bf Attention Visualization} \\
\cline{2-5}
                & {\it Src} & {\it Tgt} & {\it Src} & {\it Tgt} \\
\hline
\hline
\multirow{3}{*}{\it Good} & 
        (({\it yi}, {\it ge}), {\it zhongguo}) & {\it (to, (the, (same, china)))} & {\it \textcolor{zb_dred}{yi}~\textcolor{zb_red}{ge}~\textcolor{zb_lred}{zhongguo}} & {\it \textcolor{zb_lred}{to} \textcolor{zb_red}{the} \textcolor{zb_dred}{same china}} \\
\cline{2-5}
        & (({\it yanzhong}, {\it de}), {\it shi}) & {\it (((serious, concern), is), the)} & {\it \textcolor{zb_dred}{yanzhong}~\textcolor{zb_red}{de}~\textcolor{zb_lred}{shi}} & {\it \textcolor{zb_dred}{serious concern} \textcolor{zb_red}{is} \textcolor{zb_lred}{the}} \\
\cline{2-5}
        & ({\it jiezhi}, ({\it muqian}, {\it weizhi})) & {\it ((so, far), (this, year))} & {\it \textcolor{zb_lred}{jiezhi}~\textcolor{zb_red}{muqian}~\textcolor{zb_dred}{weizhi}} & {\it \textcolor{zb_dred}{so far} \textcolor{zb_lred}{this year}} \\
\hline
\multirow{3}{*}{\it Bad} &
        ({\it yanjin}, ({\it de}, {\it taidu})) & {\it ((be, (very, critical)), of)}  & {\it \textcolor{zb_lred}{yanjin} \textcolor{zb_red}{de} \textcolor{zb_dred}{taidu}} & {\it \textcolor{zb_red}{be} \textcolor{zb_dred}{very critical} \textcolor{zb_lred}{of}} \\
\cline{2-5}
        & ({\it zhuyao}, ({\it shi}, {\it yinwei})) & {\it ((on, (the, part)), of)} & {\it \textcolor{zb_lred}{zhuyao} \textcolor{zb_red}{shi} \textcolor{zb_dred}{yinwei}} & {\it \textcolor{zb_red}{on} \textcolor{zb_dred}{the part} \textcolor{zb_lred}{of}} \\
\cline{2-5}
        & ({\it cunzai}, ({\it de}, {\it wenti})) & {\it (problems, (of, (hong, kong)))} & {\it \textcolor{zb_lred}{cunzai} \textcolor{zb_red}{de} \textcolor{zb_dred}{wenti}} & {\it \textcolor{zb_lred}{problems} \textcolor{zb_red}{of} \textcolor{zb_dred}{hong kong}}\\
\hline
\end{tabular}
\caption{\label{attention-analysis} Examples of bilingual phrases from our translation model with both {\it phrase structures} and {\it attention visualization}. For each example, important words are highlighted in \textcolor{zb_dred}{dark red} (with the highest attention weight), \textcolor{zb_red}{red} (the second highest), \textcolor{zb_lred}{light red} (the third highest) according to their attention weights. {\it Good} = good translation pair, {\it Bad} = bad translation pair, judged according to their semantic similarity scores.}
\end{table*}

We compared BattRAE against the following three methods:
\begin{itemize}
\item
{\it Baseline}: Our baseline decoder is a state-of-the-art bracketing transduction grammar based translation system with a maximum entropy based reordering model~\cite{J97-3002,conf/acl/XiongLL06}. The features used in this baseline include: rule translation probabilities in two directions, lexical weights in two directions, target-side word number, phrase number, language model score, and the score of the maximum entropy based reordering model.
\item
{\it BRAE}: The neural model proposed by Zhang et al.~\shortcite{zhang-EtAl:2014:P14-11}. We incorporate the semantic distances computed according to BRAE as new features into the log-linear model of SMT for translation selection.
\item
{\it BCorrRAE}: The neural model proposed by Su et al.~\shortcite{su-EtAl:2015:EMNLP2} that extends BRAE with word alignment information. The structural similarities computed by BCorrRAE are integrated into the Baseline as additional features.
\end{itemize}
With respect to the two neural baselines BRAE and BCorrRAE, we used the same training data as well as the same methods as ours for hyper-parameter optimization, except for the dimensionality of word embeddings, which we set to $50$ in experiments.

Table \ref{overall-performance} summaries the experiment results of BattRAE against the other three methods on the test sets. BattRAE significantly improves translation quality on all test sets in terms of BLEU. Especially, it achieves an improvement of up to {\it 1.63} BLEU points on average over the Baseline. Comparing with the two neural baselines, our BattRAE model obtains consistent improvements on all test sets. It significantly outperforms BCorrRAE by 0.48 BLEU points, and BRAE by almost 1 BLEU point on average. We contribute these improvements to the incorporation of clues and interactions at different levels of granularity since neither BCorrRAE nor BRAE explore them.

\subsection{Attention Analysis}

Observing the significant improvements and advantages of BattRAE over BRAE and BCorrRAE, we would like to take a deeper look into how the bidimensional attention mechanism works in the BattRAE model. Specifically, we wonder which words are highly weighted by the attention mechanism. Table \ref{attention-analysis} shows some examples in our translation model. We provide phrase structures learned by RAE and visualize attention weights for these examples.

We do find that the BattRAE model is able to learn what is important for semantic similarity computation. The model can recognize the correspondence between ``{\it yige}'' and ``{\it same}'', ``{\it yanzhong}'' and ``{\it serious concern}'', ``{\it weizhi}'' and ``{\it so far}''. These word pairs tend to give high semantic similarity scores to these translation instances. In contrast, because of incorrect translation pairs ``{\it taidu}'' ({\it attitude}) vs. ``{\it very critical}'', ``{\it yinwei}'' ({\it because}) vs. ``{\it the part}'', ``{\it wenti}'' ({\it problem}) vs. ``{\it hong kong}'', the model assigns low semantic similarity scores to these negative instances. These indicate that the BattRAE model is indeed able to detect and focus on those semantically related parts of bilingual phrases.

Further observation reveals that there are strong relations between phrase structures and attention weights. Generally, the BattRAE model will assign high weights to words subsumed by many internal nodes of phrase structures. For example, we find that the correct translation of ``{\it wenti}'' actually appears in the corresponding target phrase. However, due to errors in learned phrase structures, the model fails to detect this translation. Instead, it finds an incorrect translation ``{\it hong kong}''. This suggests that the quality of learned phrase structures has an important impact on the performance of our model.

\section{Related Work}

Our work is related to {\it bilingual embeddings} and {\it attention-based neural networks}. We will introduce previous work on these two lines in this section.

\subsection{Bilingual Embeddings}

The studies on bilingual embeddings start from bilingual word embedding learning. Zou et al.~\shortcite{zou-EtAl:2013:EMNLP} use word alignments to connect embeddings of source and target words. To alleviate the reliance of bilingual embedding learning on parallel corpus, Vuli\'{c} and Moens~\shortcite{vulic-moens:2015:ACL-IJCNLP} explore document-aligned instead of sentence-aligned data, while Gouws et al.~\shortcite{conf/icml/GouwsBC15} investigate monolingual raw texts. Different from the abovementioned corpus-centered methods, Ko\v{c}isk\'{y} et al.~\shortcite{kovcisky-hermann-blunsom:2014:P14-2} develop a probabilistic model to capture deep semantic information, while Chandar et al.~\shortcite{NIPS2014_5270} testify the use of autoencoder-based methods. More recently, Luong et al.~\shortcite{luong-pham-manning:2015:VSM-NLP} jointly model context co-ocurrence information and meaning equivalent signals to learn high quality bilingual embeddings.

As phrases have long since been used as the basic translation units in SMT, bilingual phrase embeddings attract increasing interests. Since translation equivalents share the same semantic meaning, embeddings of source/target phrases can be learned with information from their counterparts. Along this line, a variety of neural models are explored: multi-layer perceptron~\cite{gao-EtAl:2014:P14-1}, RNN encoder-decoder~\cite{cho-EtAl:2014:EMNLP2014} and recursive autoencoders~\cite{zhang-EtAl:2014:P14-11,su-EtAl:2015:EMNLP2}.

The most related work to ours are the bilingual recursive autoencoders~\cite{zhang-EtAl:2014:P14-11,su-EtAl:2015:EMNLP2}. Zhang et al.~\shortcite{zhang-EtAl:2014:P14-11} represent bilingual phrases with embeddings of root nodes in bilingual RAEs, which are learned subject to transformation and distance constraints on the source and target language. Su et al.~\shortcite{su-EtAl:2015:EMNLP2} extend the model of Zhang et al.~\shortcite{zhang-EtAl:2014:P14-11} by exploring word alignments and correspondences inside source and target phrases. A major limitation of their models is that they are not able to incorporate clues of multiple levels of granularity to learn bilingual phrase embeddings, which exactly forms our basic motivation.

\subsection{Attention-Based Neural Networks}

Over the last few months, we have seen the tremendous success of attention-based neural networks in a variety of tasks, where learning alignments between different modalities is a key interest. For example, Mnih et al.~\shortcite{NIPS2014_5542} learn image objects and agent actions in a dynamic control problem. Xu et al.~\shortcite{DBLP:conf/icml/XuBKCCSZB15} exploit an attentional mechanism in the task of image caption generation. With respect to neural machine translation, Bahdanau et al.~\shortcite{DBLP:journals/corr/BahdanauCB14} succeed in jointly learning to translate and align words. Luong et al.~\shortcite{DBLP:journals/corr/LuongPM15} further evaluate different attention architectures on translation. Inspired by these works, we propose a {\it bidimensional attention network} that is suitable in the bilingual context.

In addition to the abovementioned neural models, our model is also related to the work of Socher et al.~\shortcite{NIPS2011_4204} and Yin and Sch\"{u}tze~\shortcite{yin-schutze:2015:ACL-IJCNLP} in terms of multi-granularity embeddings. The former preserves multi-granularity embeddings in tree structures and introduces a dynamic pooling technique to extract features directly from an attention matrix. The latter extends the idea of the former model to convolutional neural networks. Significantly different from their models, we introduce a bidimensional attention matrix to generate attention weights, instead of extracting features. Additionally, our attention-based model is also significantly different from the tensor networks~\cite{NIPS2013_5028} in that the attention-based model is able to handle variable-length inputs while tensor networks often assume fixed-length inputs.

We notice that the very recently proposed attentive pooling model~\cite{2016arXiv160203609D} which also aims at modeling mutual interactions between two inputs with a {\it two-way attention} mechanism that is similar to ours. The major differences between their and our work lie in the following four aspects. First, we perform a transformation ahead of attention computation in order to deal with language divergences, rather than directly compute the attention matrix. Second, we calculate attention weights via a sum-pooling approach, instead of max pooling, in order to preserve all interactions at each level of granularity. Third, we apply our bidimensional attention technique to recursive autoencoders instead of convolutional neural networks. Last, we aim at learning bilingual phrase representations rather than question answering. Most importantly, our work and theirs can be seen as two independently developed models that provide different perspectives on a new {attention mechanism}.

\section{Conclusion and Future Work}

In this paper, we have presented a bidimensional attention based recursive autoencoder to learn bilingual phrase representations. The model incorporates clues and interactions across source and target phrases at multiple levels of granularity. Through the bidimensional attention network, our model is able to integrate them into bilingual phrase embeddings. Experiment results show that our approach significantly improves translation quality.

In the future, we would like to exploit different functions to compute semantically matching scores (Eq. (\ref{attention})), and other neural models for the generation of phrase structures. Additionally, the bidimensional attention mechanism can be used in convolutional neural network and recurrent neural network. Furthermore, we are also interested in adapting our model to semantic tasks such as paraphrase identification and natural language inference.

\section*{Acknowledgements}
The authors were supported by National Natural Science Foundation of China (Grant Nos. 61303082, 61403269, 61622209 and 61672440), Natural Science Foundation of Fujian Province (Grant No. 2016J05161), and Natural Science Foundation of Jiangsu Province (Grant No. BK20140355). We also thank the anonymous reviewers for their insightful comments.

\bibliography{aaai17}
\bibliographystyle{aaai}

\end{document}